\documentclass{article}

  \usepackage[final]{neurips_2019_ml4ps}

\usepackage[utf8]{inputenc} 
\usepackage[T1]{fontenc}    
\usepackage{hyperref}       
\usepackage{url}            
\usepackage{booktabs}       
\usepackage{amsfonts}       
\usepackage{nicefrac}       
\usepackage{microtype}      

\usepackage{color}
\usepackage{amsmath}
\usepackage{amsthm}
\usepackage{graphicx}
\usepackage{algorithm}
\usepackage{algorithmic}
\usepackage{subfigure}
\usepackage{natbib}
\usepackage{float}
\usepackage{authblk}

\newcommand{\X}{\mathcal{X}}
\newcommand{\A}{\mathcal{A}}

\newcommand{\E}{\mathbb{E}}
\newcommand{\R}{\mathbb{R}}

\newtheorem{theorem}{Theorem}

\title{Offline Contextual Bayesian Optimization \\ for Nuclear Fusion}

\author[1*]{\textbf{Youngseog Chung}}
\author[1*]{\textbf{Ian Char}}
\author[1]{\textbf{Willie Neiswanger}}
\author[2]{\textbf{Kirthevasan Kandasamy}}
\author[3]{\textbf{Andrew Oakleigh Nelson}}
\author[3]{\textbf{Mark D Boyer}}
\author[3]{\textbf{Egemen Kolemen}}
\author[1]{\textbf{Jeff Schneider}}
\affil[1]{Department of Machine Learning, Carnegie Mellon University}
\affil[ ]{\textit{\{ichar, youngsec, willie, schneide\}@cs.cmu.edu}}
\affil[2]{Department of EECS, University of California Berkeley}
\affil[ ]{\textit{kandasamy@eecs.berkeley.edu}}
\affil[3]{Princeton Plasma Physics Laboratory}
\affil[ ]{\textit{\{anelson, mboyer, ekolemen\}@pppl.gov}}
\affil[*]{These authors contributed equally to this paper.}
\date{}                     
\begin{document}
\maketitle



\section{Introduction}


Nuclear fusion is regarded as the energy of the future since it presents the possibility of unlimited clean energy \citep{chen2011indispensable}. The most widespread method of realizing fusion reactions requires heating up isotopes of hydrogen to temperatures of hundreds of millions of degrees using a magnetic device called a \textit{tokamak}. In this state, the nuclei of two nearby atoms may overcome electrostatic repulsion force between them to form a single nucleus, releasing energy. In contrast to nuclear fission, this process creates much less hazardous byproducts and has a plentiful fuel source \citep{chen2011indispensable, clery2014piece}. One obstacle in utilizing fusion as a feasible energy source, however, is the stability of the reaction. At such high temperatures, the hydrogen atoms are in a plasma state, and any instability in the plasma can give rise to events called \textit{disruptions}. If disruptions occur, the plasma is quickly lost and the nuclear reaction is halted. Ideally, one would have a controller for the tokamak that makes actions in response to the current state of the plasma in order to prolong the reaction as long as possible; however, such a controller is unknown as of now.

In this work, we make preliminary steps to learning such a controller. Since learning on a real world tokamak is infeasible, we tackle this problem by attempting to learn optimal controls offline via a simulator. In our experiments, we make the simplifying assumption that the plasma will only ever be in one of eight states. Finding the best action upon seeing one of the eight states (or contexts) bares similarities to the contextual bandit setting~bandits~\citep{krause2011contextual,agrawal2013thompson, auer2002using}; however, literature usually assumes that the context is decided by nature and cannot be set by the algorithm. Intuitively speaking, being able to decide how many queries are made in each context is advantageous because some contexts may be harder to learn the optimal action for or may occur more frequently in nature. Luckily, our tokamak simulator allows for contexts to be set, and since each simulation is expensive, it is crucial that we use an algorithm that prudently picks both actions and contexts to evaluate.

In this paper we introduce an algorithm that recommends a context and action pair to evaluate at every iteration. Our algorithm falls under the umbrella of \textit{Bayesian optimization} (BO) since we introduce a prior over the reward structure of the context-action space and leverage this model for optimization. Unlike many other problems under the BO setting, our algorithm searches for an optimal action for each context rather than a single optimal action. This serves as a contrasting feature from other problems in multi-task BO \citep{multibo, toscano2018bayesian, shah2016pareto}, in which a single action that performs optimally across all objectives simultaneously is sought. Furthermore, our algorithm is theoretically grounded and can broadly be applied to "offline" problems in which context can be set explicitly. In the remainder of the paper, we briefly describe the algorithm and apply it to the nuclear fusion problem. 

\section{Algorithm}

Generally, one can think of the optimization problem for each context as a task; we are then faced with a multi-task optimization problem. Let $\X$ be the finite collection of tasks and let $\A$ be the compact set of possible actions. For our application, we assume that the set of actions is the same for each task.
Let $f: \X \times \A \rightarrow \R$ be the reward function, where $f(x, a)$ is the reward for performing action $a$ in task $x$. It is assumed that this reward function is always bounded. Let $\hat{h}: \X \rightarrow \A$ be our estimated mapping from task to action. Our goal is then to find an $\hat{h}$ which maximizes $\sum_{x \in \X} f\left(x, \hat{h}(x)\right) \omega(x)$, where $\omega(x) \geq 0$ is some weighting on $x$ (e.g. probability of seeing $x$ at evaluation time or the importance of $x$). At round $t$ of optimization, we pick a task $x_t$ and an action $a_t$ to perform a query $(x_t, a_t)$ and observe a noisy estimate of the function $y_t = f(x_t, a_t) + \epsilon_t$, where $\epsilon_t \sim N(0, \sigma_\epsilon^2)$ and is iid. Let $D_t$ be the sequence of queried tasks, actions, and rewards up to time $t$, i.e. $D_t = \{(x_1, a_1, y_1,), \ldots, (x_t, a_t, y_t)\}$. Additionally, define $\hat{y}_t(x)$ to be the best reward observed for task $x$ up to time $t$, $\hat{a}_t(x)$ to be the action made to see this corresponding reward, and $\A_t(x)$ to be the set of all actions made for task $x$ up to time $t$. In this work, we use Gaussian Processes (GP) to model the reward function. When tasks are correlated, one can use a single GP to jointly model tasks and actions; however, for this paper we only consider a fixed finite set of tasks and opt to model each task, $x$, with an independent GP with mean function $\mu_x$ and covariance function $\sigma_x$. For more information about GPs, we refer the reader to \citet{rasmussengpforml}.

Our proposed algorithm (shown in Algorithm~\ref{alg:imts}), named Multi-Task Thompson Sampling (MTS), extends the classic Thompson sampling strategy \citep{thompson1933likelihood} to the multi-task setting. The algorithm, simply put, acts optimally with respect to samples drawn from the posterior. That is, at every round a GP sample of the reward function is drawn for each task, and these samples are used as if they were the ground truth reward function to identify the task in which the most improvement can be made. After repeating this for $T$ iterations, we return the estimated mapping $\hat{h}$ such that $\hat{h}(x) = \hat{a}_T(x)$ if an evaluation was made for task $x$; if no evaluations have been made for the task, $\hat{h}(x)$ maps to an $a \in \A$ drawn uniformly at random. 

\begin{algorithm}[h!]
   \caption{Multi-Task Thompson Sampling (MTS)}
   \label{alg:imts}
\begin{algorithmic}
   \STATE {\bfseries Input:} capital $T$, initial capital $t_{init}$, mean functions $\{\mu_x\}_{x \in \X}$, kernel functions $\{\sigma_x\}_{x \in \X}$.
   \STATE Do random search on tasks in round-robin fashion until $t_{init}$ evaluations are expended.
   \FOR{$t=t_{init} + 1$ {\bfseries to} $T$}
   \STATE Draw $\widetilde{f}(x, \cdot) \sim GP(\mu_x, \sigma_x) | D_{t - 1} \quad \forall x \in \X$.
   \STATE Set $x_t = \underset{x \in \X}{\textrm{argmax}} \left[\left(\max_{a \in \A}\widetilde{f}(x, a) - \max_{a \in \A_t(x)}\widetilde{f}(x, a) \right) \omega(x)\right]$.
   \STATE Set $a_t = \underset{a \in \A}{\textrm{argmax}} \widetilde{f}(x_t, a)$.
   \STATE Observe $y_t = f(x_t, a_t)$.
   \STATE Update $D_t = D_{t - 1} \cup \{(x_t, a_t, y_t)\}$.
   \ENDFOR
   \STATE {\bfseries Output:} $\hat{h}$
\end{algorithmic}
\end{algorithm}

One benefit of this algorithm is that it comes with theoretic guarantees. For the following, define $a_t^*(x)$ to be the past action played for task $x$ up to time $t$ that yields the largest expected reward.  That is,
\begin{align*}
    a_t^*(x) := \begin{cases}
        \underset{a \in \A_t(x)}{\textrm{argmax}} f(x, a) & \mathcal{A}_t(x) \neq \emptyset \\
        \underset{a \in \A}{\textrm{argmin}} f(x, a) & \textrm{else}
    \end{cases}
\end{align*}
Note that $a_t^*(x)$ has an implicit dependence on $f$.

\begin{theorem}\label{thm:regret}
Define the maximum information gain to be $\gamma_T := \max_{D_T} I(D_T;f)$, where $I(\cdot; \cdot)$ is the Shannon mutual information. Assume that $\X$ and $\A$ are finite. Then if Algorithm~\ref{alg:imts} is played for $T$ rounds where $t_{init} = 0$ 
\begin{align*}
    \E\left[R_{T, f}\right] \leq |\X| \left( \frac{1}{T} + \sqrt{\frac{|\X| |\A| \gamma_T}{2T}} \right)
\end{align*}
where the expectation is with respect to the data sequence collected and $f$, and where $R_{T, f}$ is defined to be
\begin{align*}
    R_{T, f} := \frac{\sum_{x \in \X} \omega(x) \left( \max_{a \in \A} f(x, a) - f(x, a_T^*(x)) \right)}{\sum_{x \in \X} \omega(x) \left( \max_{a \in \A} f(x, a) - \min_{a \in \A} f(x, a) \right)}
\end{align*}
when the denominator is not $0$. Otherwise, $R_{T, f}$ takes the value of $0$.
\end{theorem}

The proof relies on ideas from \citet{kandasamy2019myopic}, and the details can be found in Appendix A of \citet{char2019offline}. The result gives a bound on the normalized simple regret summed across tasks where the the $\sqrt{|\X||\A|}$ factor in the theorem accounts for the number of actions that can be taken at every step, and the $\sqrt{\gamma_T}$ factor characterizes the complexity of the prior over the tasks. Intuitively, this result shows that there is no task in which we will have especially bad results, and when $\gamma_T = o(T)$, the normalized simple regret converges to 0 in expectation for every task. Finally, we note that these types of results can usually be generalized to infinite action spaces via known techniques \citep{russo2016information, bubeck2011x} 

\section{Application to Nuclear Fusion} \label{sec:fusion}

\subsection{Tokamak Simulator (TRANSP) and Overview of Problem} \label{app:fusion_prelim}

We use the TRANSP program \citep{grierson2018orchestrating} to simulate fusion reactions on the DIII-D tokamak, a tokamak in San Diego that is operated by General Atomics. TRANSP is a time-dependent transport code used for interpretive analysis and predictive simulations of tokamaks. Access to TRANSP and running TRANSP experiments were possible thanks to our collaborators at Princeton Plasma Physics Lab. TRANSP operates by simulating real-world experiments (referred to as "shots") that were conducted on DIII-D. By running the predictive module of TRANSP, we are able to predict how changes in controls would affect the plasma. When simulating a given shot (a simulation on TRANSP is referred to as a "run"), we can identify variables at each time step that correspond to the state of the plasma. One such variable that we focus on is $\beta_n$, which is a ratio of the pressure of the plasma to the magnetic energy density. $\beta_n$ serves as a proxy for the economic output of the reaction. Besides this quantity, we also consider the \textit{total energy eigenvalues}, which represent the amount of change in energy within and outside the plasma due to certain perturbations. In particular, we focus on the minimum value of the total energy eigenvalues, which we will refer to as $\Delta \omega$. $\Delta \omega$ serves as a proxy for the stability of the plasma. When conducting a simulation, we apply controls that specify parameters of the neutral beams, which include power, energy, full energy fraction and half energy fraction. The DIII-D tokamak has a total of 8 neutral beams, 6 of which are co-current beams (inject in the same direction as the plasma current) and 2 of which are counter-current beams (inject in the opposite direction of the plasma current). In our experiments, we confine the action space to 2 dimensions: power coefficient of co-current beams and counter-current beams, each with domain [0.001, 1.0]. These power coefficients are applied by multiplying the maximum power of the set of beams by the coefficient. By ranging the power coefficient from 0.001 to 1.0, we essentially scale the beam powers from the minimum to the maximum power level possible. 

In our experiments, we consider 8 distinct states of the plasma, which are represented by 8 shots. In all 8 shots, a common instability called \textit{tearing} occurred. 
Ideally, we would like to perform preventative measures once we sense a tear is about to occur. Therefore, we start the simulation 150 ms before time of tearing and run the simulator until 150 ms after the tearing. After the run completes, we extract the $\Delta \omega$ and $\beta_n$ values at 5ms increments throughout the duration of the run (total 300 ms) and average them to produce $\overline{\Delta \omega}$ and $\overline{\beta}_n$. In order to balance between stability and the pressure in the tokamak, we set our reward to be $10\overline{\beta}_n + 100\overline{\Delta \omega}$, where we chose coefficients based on the scales of each value to make the two objective components roughly equal. In summary, we optimize a combination of pressure and stability of the plasma, for each of the 8 different plasma states (8 tasks or contexts) simultaneously, by changing the power level of the co-current and counter-current beams (2D controls). 

\subsection{Tokamak Control Optimization} \label{sec:fusion_imts}

\begin{figure*}
    \centering
    \subfigure[]{\includegraphics[width=.33\linewidth, height=0.20\textheight]{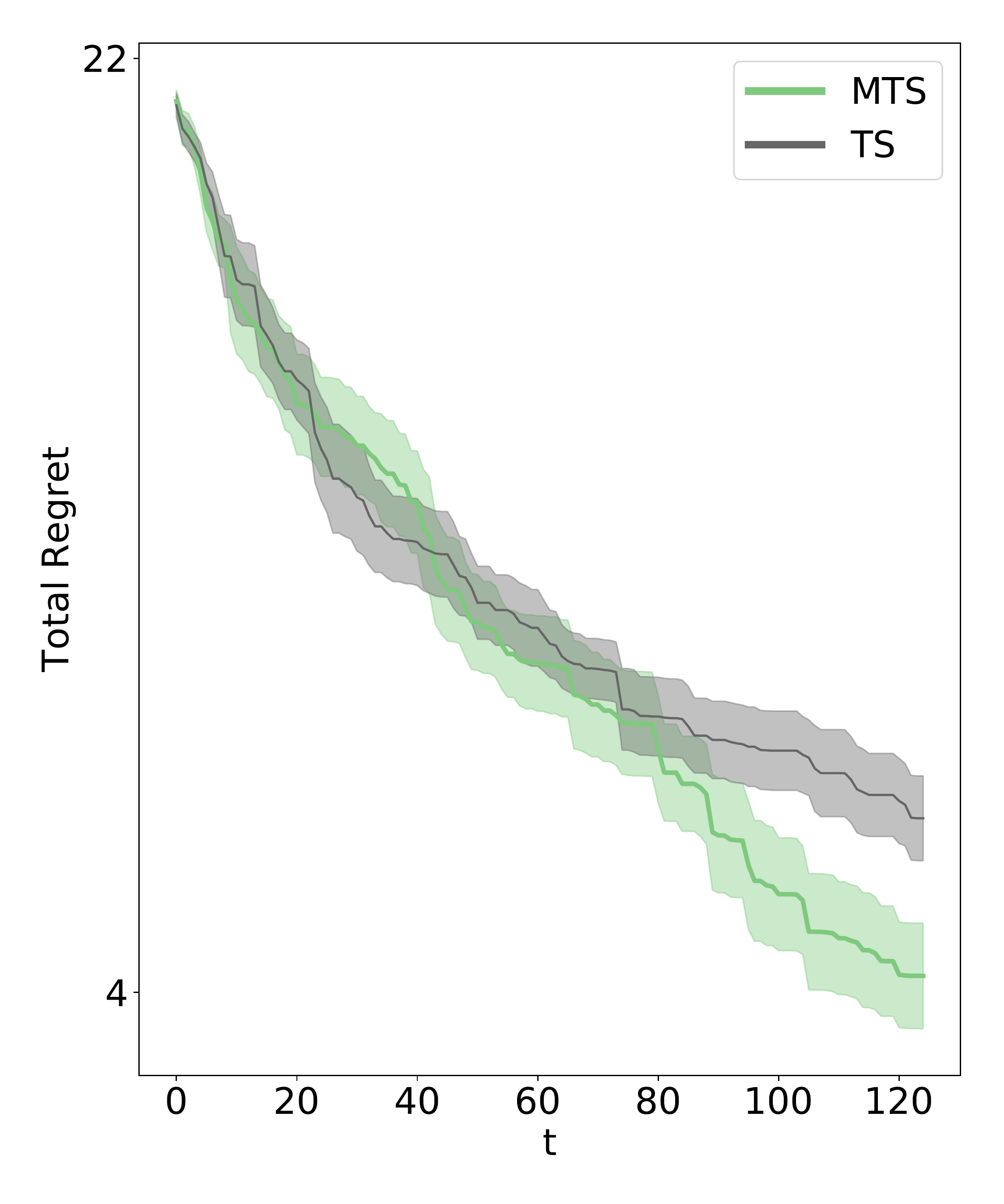}}
    \subfigure[]{\includegraphics[width=.66\linewidth, height=0.20\textheight]{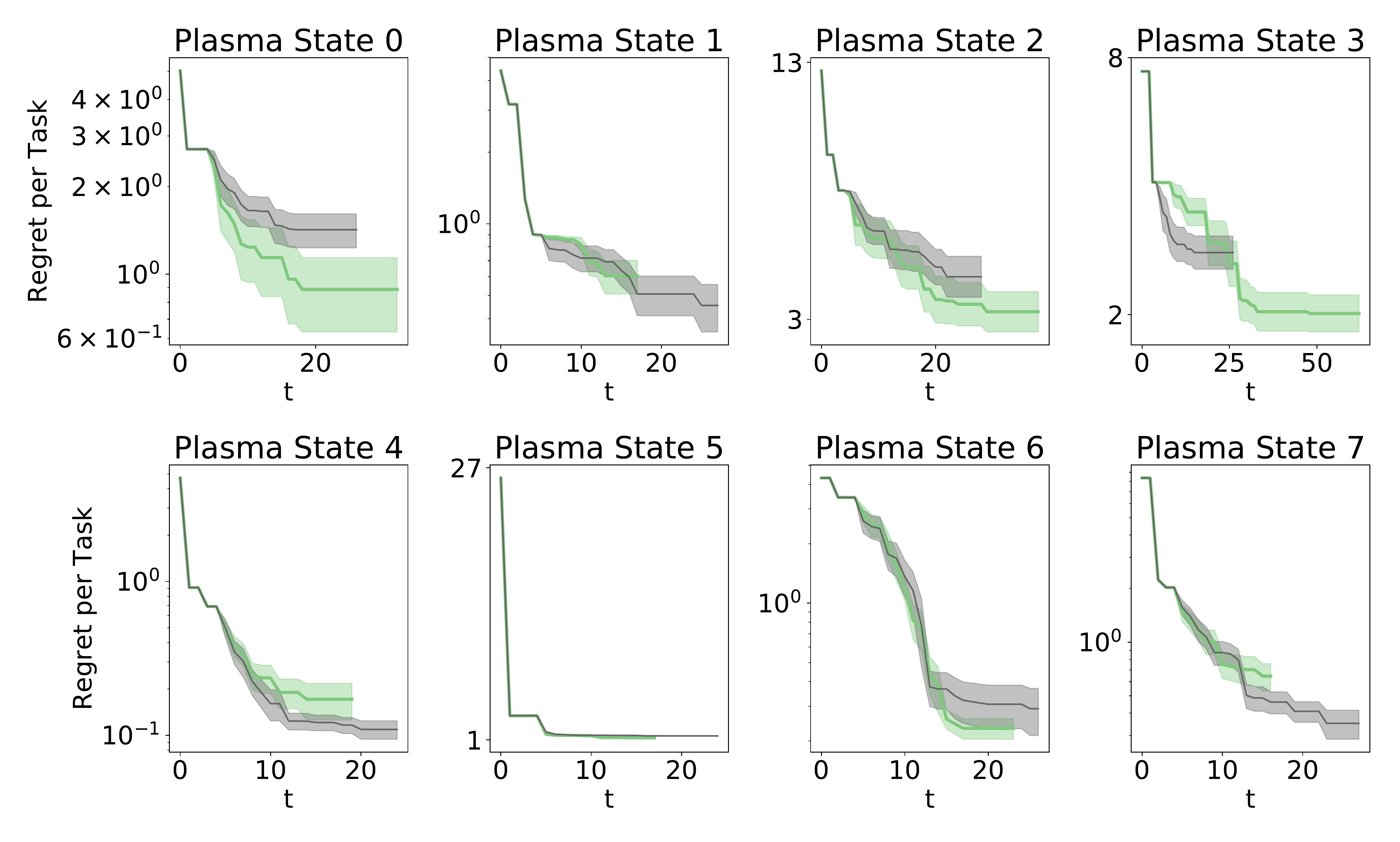}}
    \caption{\textbf{Fusion Simulation Experiments.} Each of the above show average values and standard error from 10 trials. \textbf{(a)} shows the log total regret summed across all states and \textbf{(b)} log regret achieved in each state. Note that curves differ in length for (b) since different amounts of resources were allocated for each state. Note that regret was approximated by treating the optimum as the maximum value observed for each state, plus a small $\epsilon$ term.}
    \label{fig:fusion_reward}
\end{figure*}

We optimize the reward produced from TRANSP simulations using both MTS and a method that chooses shots (or plasma states) uniformly at random then performs the standard Thompson sampling procedure. The optimization experiment results presented in Figure~\ref{fig:fusion_reward} are averaged over 10 trials, each with 125 query capital. In each trial and for each shot, 5 initial points are drawn uniformly at random for evaluation to form an initial GP. Each task is modeled by an independent GP with an RBF kernel and hyperparameters are tuned for each GP every time an observation is seen for its corresponding shot by maximizing the marginal likelihood. We parallelized the experiments by having a batch of 20 workers, each making evaluations according to the respective algorithms using a shared pool of collected data. This process is suboptimal since workers operate asynchronously (i.e. they do not wait to see the data other workers will collect); however, \citet{kandasamy2018parallelised} showed that this approach is not unreasonable for the standard Thompson sampling setting. The results demonstrate how MTS is able to achieve better performance by focusing its resources intelligently. Looking at Figure~\ref{fig:fusion_reward} (b), in contexts where reward (and hence regret) levels off quickly (e.g. plasma states 4 and 5), MTS is able to recognize that resources should be allocated in other contexts where higher improvement is expected. This is reflected with more queries and better optimization in plasma states 2 and 3.

\subsection{Discussion of Results and Future Directions}

\begin{figure*}
    \centering
    \subfigure[]{\includegraphics[width=.32\linewidth, height=0.13\textheight]{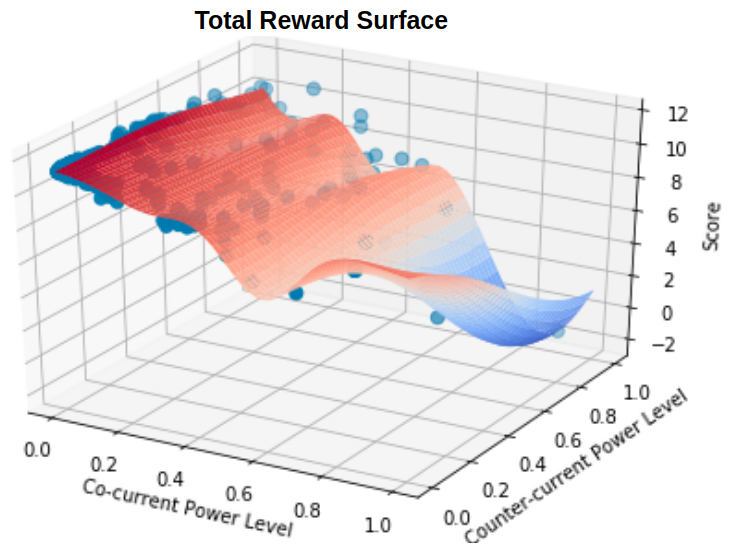}}
    \subfigure[]{\includegraphics[width=.32\linewidth, height=0.13\textheight]{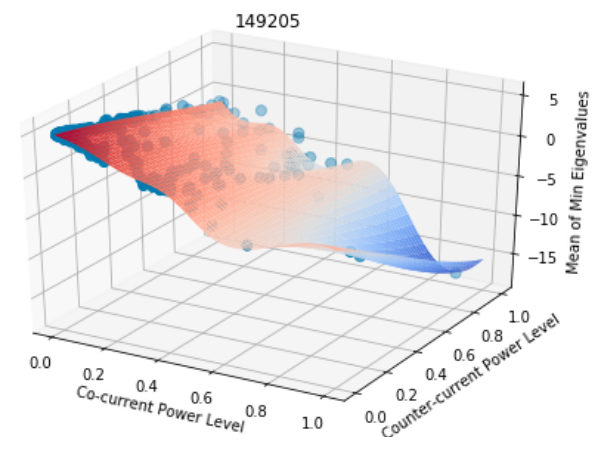}}
    \subfigure[]{\includegraphics[width=.32\linewidth, height=0.13\textheight]{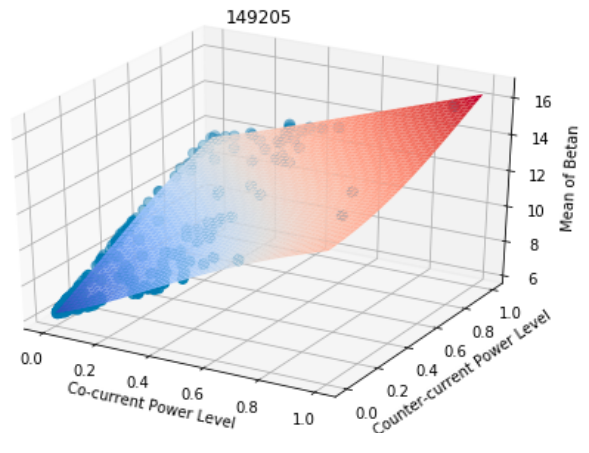}}
    \caption{\textbf{Reward and Reward Component Surfaces.} The surfaces have been estimated by fitting a GP to the queried points from shot 149205, which corresponds to plasma state 1. Each point in the space shows a value returned by the simulator.}
    \label{fig:surfaces}
\end{figure*}

Figure~\ref{fig:surfaces} shows the total reward and reward component surfaces for one of the 8 shots from the experiments conducted in Section~\ref{sec:fusion_imts}. From the total reward surface (Figure~\ref{fig:surfaces} (a)), we can see that beam power should be scaled down for maximum total reward. However, the surfaces of each reward component indicate the negative correlation between plasma stability (Figure~\ref{fig:surfaces} (b)) and pressure (Figure~\ref{fig:surfaces} (c)): as beam power is scaled up, stability increases but pressure decreases. Hence, there is a fine balance in optimizing a combination of these components, which is dependent on the weighting between the two objectives. This raises further questions about exactly what weighting would be optimal for actual plasma behavior over a longer period of time. In addition, these results provide interesting insights to the fusion community. Although there has been some previous work applying machine learning to fusion in the past \citep{cannas2013automatic, tang2016big, montes2019machine, kates2019predicting, baltz2017achievement}, to the best of our knowledge this is the first work towards learning a tokamak controller offline with no human intervention.


In the future, we hope to discover more interesting results by increasing our action space and forming different reward functions. We are also working to expand this work to finding a policy over a continuous set of plasma states, rather than just a subset of eight. From an algorithmic standpoint, this problem has been explored by \citet{ginsbourger2014bayesian} and \citet{pearce2018continuous}. We have preliminary evidence showing that a variation of our algorithm performs competitively with theirs. Lastly, readers may remark that the problem of tokamak control is actually a reinforcement learning problem, since we should be searching for an optimal policy that makes a sequence of actions rather than a single action. Because the simulator is expensive (it takes approximately 2 hours to simulate one control evaluation lasting 300 ms), we chose to limit the scope of the problem in this work; however, we wish to revisit this in the future.

\section{Acknowledgements}

This material is based upon work supported by the National Science Foundation Graduate Research Fellowship Program under Grant No. DGE1252522 and DGE1745016. Willie Neiswanger is also supported by NSF grants CCF1629559 and IIS1563887. Any opinions, findings, and conclusions or recommendations expressed in this material are those of the authors and do not necessarily reflect the views of the National Science Foundation.

Youngseog Chung is supported by the Kwanjeong Educational Foundation.

The authors would also like to thank the reviewers of the \textit{Machine Learning and the Physical Sciences} workshop for their helpful feedback.

\bibliography{camera_ready}
\bibliographystyle{plainnat}

\end{document}